\documentclass{bmvc2k}
\usepackage{algorithm} 
\usepackage{algpseudocode} 
\usepackage{amsfonts}
\usepackage{svg}
\usepackage{adjustbox}
\usepackage{wrapfig}
\usepackage{caption}
\usepackage{parskip}
\usepackage{float} % for the [H] option
\usepackage{booktabs} % for the \toprule, \midrule, and \bottomrule 
\usepackage{caption} 
\usepackage{multirow}
\usepackage{array}
\usepackage{graphicx}

\usepackage{amssymb}
\usepackage{pifont}
\usepackage{amsmath}
\usepackage{color}
\usepackage{listings}
\usepackage{xcolor}
\usepackage{graphicx}
\usepackage[most]{tcolorbox}

\definecolor{codegreen}{rgb}{0,0.6,0}
\definecolor{codegray}{rgb}{0.5,0.5,0.5}
\definecolor{codepurple}{rgb}{0.58,0,0.82}
\definecolor{backcolour}{rgb}{1,1,1}

\lstdefinestyle{mystyle}{
    backgroundcolor=\color{backcolour},   
    commentstyle=\color{codegreen},
    keywordstyle=\color{magenta},
    numberstyle=\tiny\color{codegray},
    stringstyle=\color{codepurple},
    basicstyle=\ttfamily\footnotesize,
    breakatwhitespace=false,         
    breaklines=true,                 
    captionpos=b,                    
    keepspaces=true,                 
    numbers=left,                    
    numbersep=5pt,                  
    showspaces=false,                
    showstringspaces=false,
    showtabs=false,                  
    tabsize=2
}

\title{CSAD: Unsupervised Component Segmentation for Logical Anomaly Detection}

\addauthor{Yu Hsuan Hsieh}{ss111062646@gapp.nthu.edu.tw}{1}
\addauthor{Shang Hong Lai}{lai@cs.nthu.edu.tw}{1}
\addinstitution{
 National Tsing Hua University \\
 Hsinchu, Taiwan
}

\runninghead{Y.H. Hsieh, S.H. Lai}{CSAD:Unsupervised component Segmentation}

\begin{document}
\setlength\extrarowheight{1pt}
\maketitle
\captionsetup{skip=5pt}
\captionsetup[table]{skip=5pt}
\captionsetup{font={color=bmv@captioncolor}}
\begin{abstract}

To improve logical anomaly detection, some previous works have integrated segmentation techniques with conventional anomaly detection methods. Although these methods are effective, they frequently lead to unsatisfactory segmentation results and require manual annotations. To address these drawbacks, we develop an unsupervised component segmentation technique that leverages foundation models to autonomously generate training labels for a lightweight segmentation network without human labeling. Integrating this new segmentation technique with our proposed Patch Histogram module and the Local-Global Student-Teacher (LGST) module, we achieve a detection AUROC of 95.3\% in the MVTec LOCO AD dataset, which surpasses previous SOTA methods. Furthermore, our proposed method provides lower latency and higher throughput than most existing approaches.

\end{abstract}

%-------------------------------------------------------------------------

\section{Introduction}

\begin{wrapfigure}[16]{r}{0.55\textwidth}
\vspace{-30pt}
\includegraphics[width=0.55\textwidth]{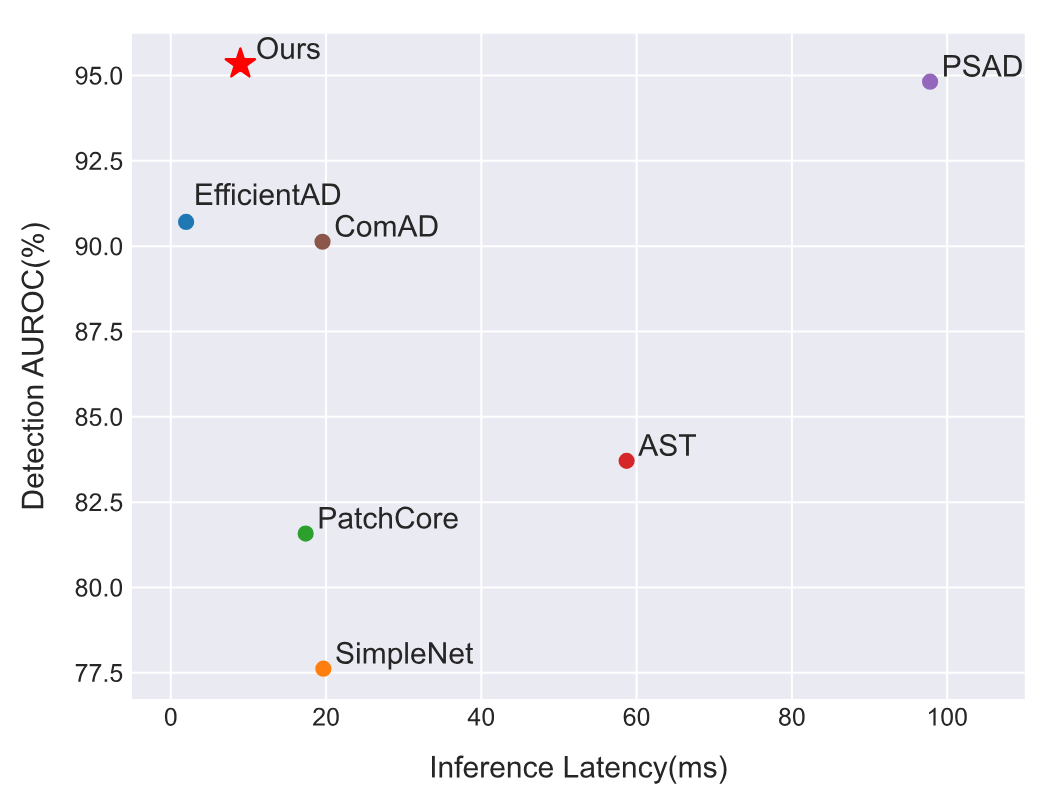}
  \caption{The speed-performance plot on the MVTec LOCO AD benchmark. The x- and y-axis indicate inference latency and average detection AUROC, respectively.}
  \label{fig:speed}
\end{wrapfigure}
In industrial anomaly detection (AD), previous works~\cite{batzner2024efficientad,liu2023simplenet,rudolph2023asymmetric,zavrtanik2021draem,tien2023revisiting} have demonstrated excellent performance in various datasets, like MVTec AD~\cite{8954181} dataset and VisA~\cite{zou2022spot} dataset. However, the research on logical anomaly detection remains underdeveloped. In this anomaly detection task, we focus on identifying violations of underlying logical constraints in images, such as incorrect quantities, arrangements, and combinations of object components. Although previous efforts~\cite{LIU2023102161,kim2024few} have achieved progress by segmenting components and calculating their areas or quantities, they struggle to recognize components with similar textures or even require manual annotations.

\quad In the field of image segmentation, foundation models, such as the Segment Anything Model (SAM)~\cite{kirillov2023segany} and Grounding DINO ~\cite{liu2023grounding}, are making notable strides. SAM excels in isolating objects within images using points or boxes as prompts. However, this model cannot identify objects from general text descriptions.
Grounding DINO is an open-set object detection model that detects objects based on text prompts and is trained on a vast dataset that includes detection and visual grounding data. Although it has a strong zero-shot object detection performance, it requires textual input.
By combining these models, Grounded-SAM~\cite{ren2024grounded} can segment any semantic object in the image via text prompts in the wild with high segmentation quality.
However, direct application of these models to industrial images does not yield satisfactory results~\cite{kim2024few}, as these methods falter when encountering objects absent during the training phase, significantly impeding their applicability of component segmentation.
Furthermore, certain industrial products, such as long and short screws in the "screw bag" category, cannot be consistently segmented using only the term "screw."
To address this challenge, we exploit multiple foundation models to generate semantic pseudo-labels of object components and train a segmentation network in a semi-supervised setting. Training a lightweight segmentation network allows efficient segmentation of industrial images without relying on heavy foundation models in the inference stage.
Our contributions can be
summarized as follows:

\begin{itemize}
\itemsep-0.2em
\leftmargini=6mm
    \item We develop an unsupervised method to generate semantic pseudo-label maps for training a lightweight component segmentation model for a specific logical anomaly detection task without human labeling. 
    \item We propose a Patch Histogram module based on an unsupervised image segmentation network trained from semantic pseudo-labels that can effectively detect both positional and quantity abnormalities of the components in an image.
    \item We develop a Local-Global Student-Teacher(LGST) module to detect both small- and large-scale anomalies.
    \item Our approach achieves state-of-the-art performance in identifying logical and structural anomalies, with lower latency and higher throughput than most existing methods.
    
\end{itemize}

\section{Related Work}
\subsection{Conventional AD methods}
Recently, anomaly detection has predominantly followed the unsupervised setting, wherein the model is trained solely with normal samples and is tested against normal and abnormal samples. Some works~\cite{cohen2020sub,defard2021padim,huang2022regad} model deep features with multivariate Gaussian distributions and compute the Mahalanobis distance to the training set as the anomaly score. Using kNN-based anomaly detection on deep features extracted by a pretrained neural network~\cite{roth2022towards,bae2023pni,mcintosh2023inter,hyun2024reconpatch}, this approach offers a significant advantage as it does not require training. Student-teacher-based (ST-based) methods~\cite{Bergmann_2020,deng2022anomaly,tien2023revisiting,gu2023remembering,Wang_2021_CVPR} were developed based on the assumption that a student model, trained on normal samples only, will exhibit a different feature distribution in anomalous regions compared to a pretrained teacher model. 

\subsection{Logical AD methods}
THFR~\cite{guo2023template} is an ST-based method that employs bottleneck compression and utilizes normal images as templates to preserve and restore features in anomalous images. 
DSKD~\cite{zhang2024contextual} constructs two reverse knowledge distillation models: the local student, which reconstructs low-level features to identify structural anomalies, and the global student, which leverages global context to detect logical anomalies. 
EfficientAD~\cite{batzner2024efficientad} implements a knowledge distillation framework composed of a teacher distilled from a pretrained encoder, a student, and an autoencoder. Both the student-teacher and autoencoder-student pairs are specifically designed to detect small and large-scale anomalies, respectively. Both of the following methods utilize segmentation to improve the performance in detecting logical anomalies. ComAD~\cite{LIU2023102161} introduces an unsupervised segmentation method using DINO~\cite{caron2021emerging} as a feature extractor. This method segments images by clustering the features, models normal features of object components with a memory bank, and compares the region features of test samples with those of training samples to detect logical anomalies effectively. 
PSAD~\cite{kim2024few} trains a segmentation network in a semi-supervised setting using human-annotated ground-truth label maps. It constructs three memory banks: a standard PatchCore~\cite{roth2022towards} patch feature memory bank, a class histogram memory bank that stores class distribution histograms, and a class composition memory bank that aggregates class embeddings. The anomaly detection is then performed by aggregating the anomaly scores from these three memory banks.

\section{Methodology}
\subsection{Semantic Pseudo-label Generation}

\noindent
{\bf Component-level Segmentation}\quad 

\begin{wrapfigure}[11]{r}{0.55\textwidth}
\vspace{-20pt}
\includegraphics[width=0.55\textwidth]{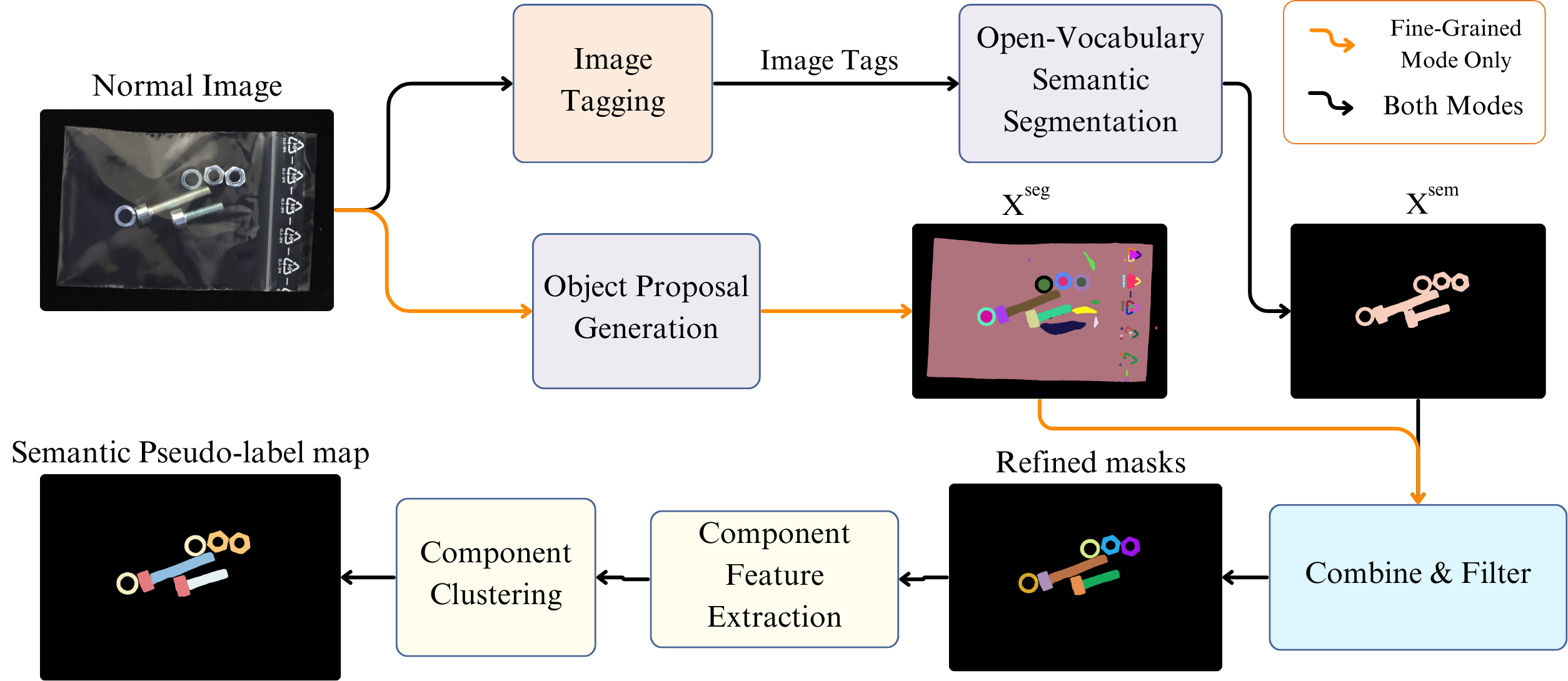}
  \caption{Proposed semantic pseudo-label generation that generates semantic pseudo-labels from normal images only.}
  \label{fig:labelgen}
\end{wrapfigure}

A primary challenge with unsupervised segmentation is the tendency toward under- or over-segmentation.
To address this problem, we implement two modes in the semantic pseudo-label generation process: fine-grained and coarse-grained modes. Typically, the coarse-grained mode generates enough semantic labels for logical anomaly detection. However, the fine-grained mode is employed when object components are too complex to discern using simple image tags. In our experiments, we use the fine-grained mode for the "screw bag" and the "juice bottle," while the coarse-grained mode is used for other categories. Here, components of the same class are defined as segments sharing similar visual features.

\quad Figure \ref{fig:labelgen} shows the procedure of our semantic pseudo-label generation. To create a semantic-aware component segmentation, we initially employ an open-vocabulary semantic segmentation model and utilize an object proposal generation model to segment each component within the image more precisely. 

\noindent
{\bf Image Tags Generation}\quad We utilize an image-tagging model, Recognize Anything++\\ (RAM++) \cite{huang2023open} model, to automatically generate text prompts for semantic segmentation. RAM++, an advanced iteration of the original Recognize Anything(RAM)~\cite{zhang2023recognize} image tagging foundation model, is trained for the open-set image tagging challenge. It operates by using a normal image from each category as input. Subsequently, the generated tags are manually refined to remove non-noun or background tags such as "attach" and "connect" found in the "splicing connector" category and "zip-lock bag" from "screw bag." A list of tags for all categories is available in the supplementary material.

\noindent
{\bf Mask Refinement}\quad Next, for a training image $X_i$ , we use the text prompts previously created and Grounded-SAM as our open-vocabulary semantic segmentation model to generate a semantic segmentation map $X^{sem}_i$ . Additionally, we employ the automatic mask generation feature of the SAM model as our object proposal generation model to produce all possible masks of components, denoted by $X^{seg}_i$, allowing segmentation of each object component. In the categories operating under the fine-grained mode, an algorithm is implemented to filter noise in $X^{seg}_i$ using $X^{sem}_i$, producing refined masks $X^{ref}_i$. A detailed description of this algorithm can be found in the supplementary material. The refined masks are identical to $X^{sem}_i$ for categories operating in the coarse-grained mode.

\noindent
{\bf Component Feature Extraction}\quad After obtaining the refined masks, we need to cluster them into semantic pseudo-labels by their visual features. Since we expect the components with the same shape and texture to belong to the same cluster, we need to eliminate the influence of rotation on an object component. To achieve this, we crop each segment and resize the diagonal of the minimum bounding rectangle of the segment to 64x64. After that, we apply rotation augmentation on each component and extract their corresponding features from the fourth layer of a pretrained CNN, with the rotation-augmented feature map denoted by $f^{rot} \in \mathbb{R}^{R \times C \times H \times W}, $ where $R$ is the number of rotations, and $C$, $H$, $W$ represent feature channel, height, and width, respectively. We take the average of the feature maps over H, W, and R dimensions to form a rotation-invariant feature vector $f^{com} \in \mathbb{R}^{C}$ representing the component.

\noindent
{\bf Component Clustering}\quad We cluster all components in the training samples by their $f^{com}$ using MeanShift~\cite{comaniciu2002mean} to avoid the cluster number selection. Assuming that each component should appear in all images, we discard any clusters with less than $\alpha$ members. We denote the number of these filtered clusters as $N_{cls}$, and the semantic pseudo-label map $Y_i$ with each pixel value representing its class number. In our experiments, we defined $\alpha$ as half of the training samples.

\noindent
{\bf Filtering Unreliable Semantic Pseudo-label Maps}\quad To ensure greater consistency across semantic pseudo-label maps, we compute a class histogram $ Hist(Y_i) \in \mathbb{R}^{N_{cls}} $  as described in Equation \eqref{eq1}
\begin{equation}
Hist(Y_i)[k]=\frac{1}{W\times H} \sum_{x=1}^{W}\sum_{y=1}^{H}\mathbb{I} (Y_i(x,y)=k), \text{ where } k = 1, 2, \ldots, N_{cls}
\label{eq1}
\end{equation}
 where $\mathbb{I} (Y_i(x,y)=k)$ is an indicator function that equals 1 if $Y_i(x,y)=k$ and 0 otherwise. Assuming that the size of the object components remains consistent across normal images, the corresponding class histogram should be similar. We apply HDBSCAN~\cite{campello2013density}, an enhanced version of  DBSCAN~\cite{ester1996density}, to the histograms to eliminate incorrect semantic pseudo-label maps, discarding those that do not belong to the largest cluster.

\quad To this end, we obtain $N_l$ normal images with high-quality semantic pseudo-label maps (labeled images), denoted as $\{X^{l}_1,\;...,\;X^{l}_{N_l}\}$, the corresponding semantic pseudo-label maps are $\{Y^{l}_1,\;...,\;Y^{l}_{N_l}\}$, and the rest of the unlabeled images are $\{X^{u}_1,\;...,\;X^{u}_{N_u}\}$.

\subsection{Component Segmentation}

\begin{figure}[H]
\captionsetup{width=\linewidth}
    \centering
    \includegraphics[width=0.5\textwidth]{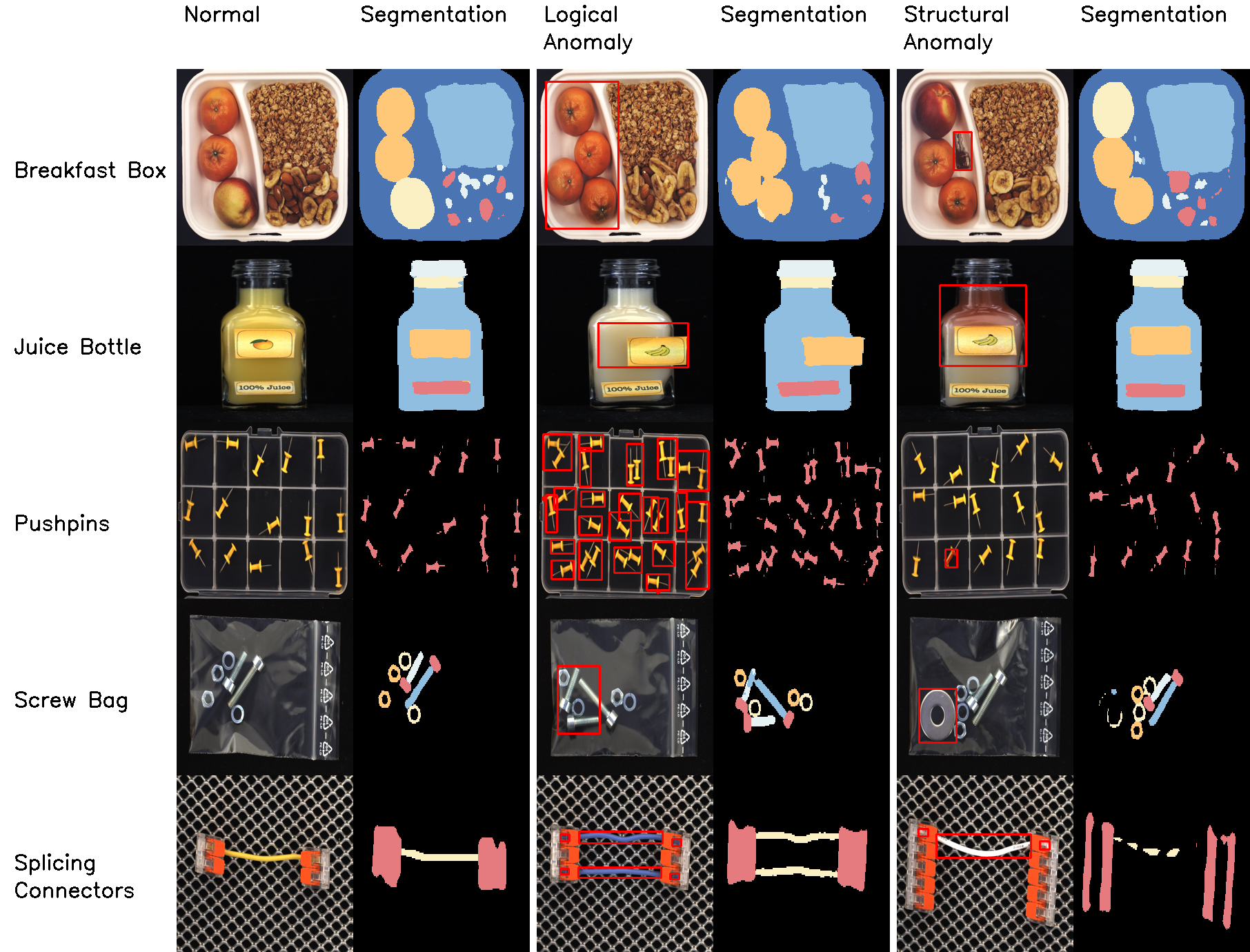}
    \caption{Segmentation result of five categories from MVTec LOCO AD. The red bounding box indicates the anomalous region of the image, and the color in the segmentation image represents the class label of the pixel.}
    \label{fig:seg_vis}
\end{figure}

{\bf Segmentation Network Architecture}\quad We utilize a DeepLabV3+~\cite{chen2018encoder} decoder that processes feature maps of multiple scales drawn from the intermediate layers of a pretrained CNN. The overall architecture of the segmentation network is a U-shape CNN with skip connections. Example images of the segmentation result are shown in Figure \ref{fig:seg_vis}.

\noindent
{\bf Logical Synthetic Anomalies (LSA)}\quad 
Training with normal images only for component segmentation may cause the model to overfit to component locations. In other words, it may predict labels based on the component's position rather than its semantic features. 
To address this problem, we propose Logical Synthetic Anomalies (LSA). This augmentation places additional components into the image to simulate logical anomalies, such as incorrect component counts or misplacements. In this way, we can increase the diversity of the training set for component segmentation, especially since the segmentation network relies only on a subset of normal images for supervised training. Example images can be found in the supplementary material.

\noindent
{\bf Loss Functions}\quad
For the training of the segmentation model, we employ Cross Entropy loss $\mathcal{L}_{CE}$, Dice loss $\mathcal{L}_{Dice}$~\cite{milletari2016v}, and Focal loss $\mathcal{L}_{Focal}$ ~\cite{lin2017focal} for labeled images. Following the approach described in PSAD, histogram matching loss $\mathcal{L}_{Hist}$ is applied to unlabeled images. The loss measures the difference between the class histogram of the input image and that of a randomly sampled labeled image, i.e., 
\begin{equation}
\mathcal{L}_{hist} = \frac{1}{N_{cls}} \sum_{k=1}^{N_{cls}} \left \| Hist(Y_i)[k] - Hist(P_i)[k] \right \|
\label{eq2}
\end{equation}
where $P_i$ is the label map predicted by the segmentation network from $X_i$. Furthermore, the entropy loss $\mathcal{L}_{Ent}$~\cite{wang2022semi} is used to reduce the uncertainty of the prediction. The total loss for training the segmentation network model is given in Equation \eqref{eq3}.
\begin{equation}
\mathcal{L}_{seg} = \lambda_1\:\mathcal{L}_{CE}+\lambda_2\:\mathcal{L}_{Dice}+\lambda_3\:\mathcal{L}_{Focal}+\lambda_4\:\mathcal{L}_{Hist}+\lambda_5\:\mathcal{L}_{Ent}
\label{eq3}
\end{equation}

\subsection{Patch Histogram}
{\bf Class Histogram}\quad 
Using the segmentation map $P_i$ predicted by the segmentation network, we can identify logical anomalies by comparing the class histogram of a given image with that of normal images.
By calculating the Mahalanobis distance from the class histogram of the testing sample to the class histograms of the training samples, we obtain the class histogram anomaly score as follows:
\begin{equation}
M(P_i)= \sqrt{(Hist(P_i)-\mu_{hist})^T\textstyle\sum_{hist}^{-1}(Hist(P_i)-\mu_{hist})}
\label{eq4}
\end{equation}
where $\mu_{hist}$ and $\sum_{hist}$ represent the mean and covariance matrix of class histograms estimated from training samples.

\begin{wrapfigure}{r}{0.5\textwidth}
  \vspace{5pt}
  \includegraphics[width=0.5\textwidth]{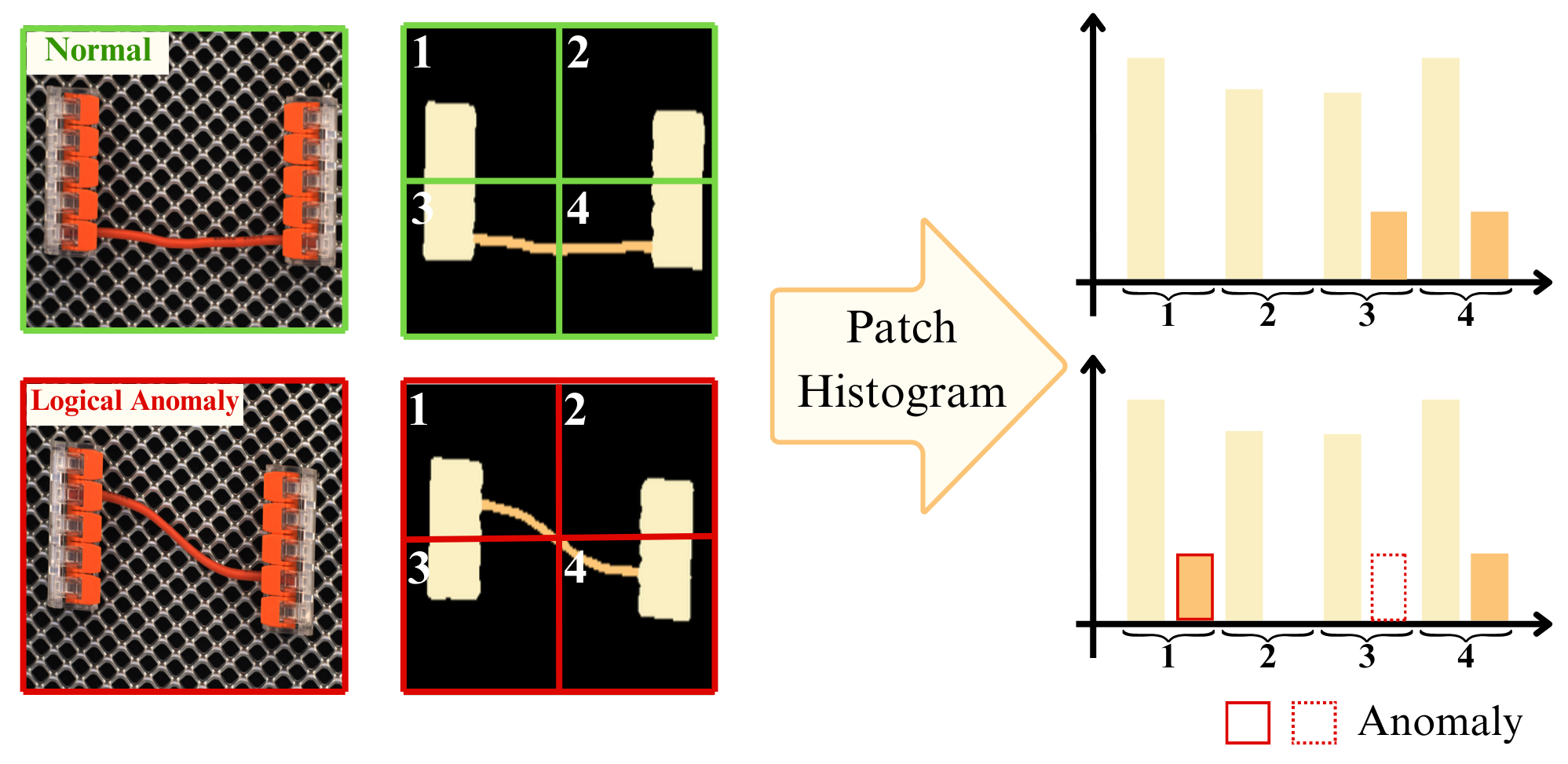}
  \caption{An example illustrating the effectiveness of patch histograms in addressing position-related logical anomalies.}
  \label{fig:patchhist}
\end{wrapfigure}

\noindent
{\bf Patch Histogram}
\quad By analyzing the class histogram of the image, we can identify logical anomalies, including incorrect quantities of object components and some structural inconsistencies. However, some logical anomalies cannot be detected through class histograms. For instance, as illustrated in Figure \ref{fig:patchhist}, in "splicing connectors," a type of logical anomaly is "Wrong\_cable\_location," where a wire is connected to an incorrect socket. This type of anomaly cannot be identified by the class histogram, as the distribution of different component classes remains consistent with that of a normal image. 

\quad To address this limitation, we introduce the patch histogram. This technique divides the predicted segmentation map into a $P \times P$ grid, constructs a class histogram for each grid cell, and concatenates these histograms to form a patch histogram vector $H^p_i \in \mathbb{R}^{(Cls\times P\times P)}$. The calculation of the anomaly score follows the same method as Equation \eqref{eq2}.

\subsection{Local-Global Student-Teacher(LGST)}
\begin{figure}
    \centering
    \includegraphics[width=\textwidth]{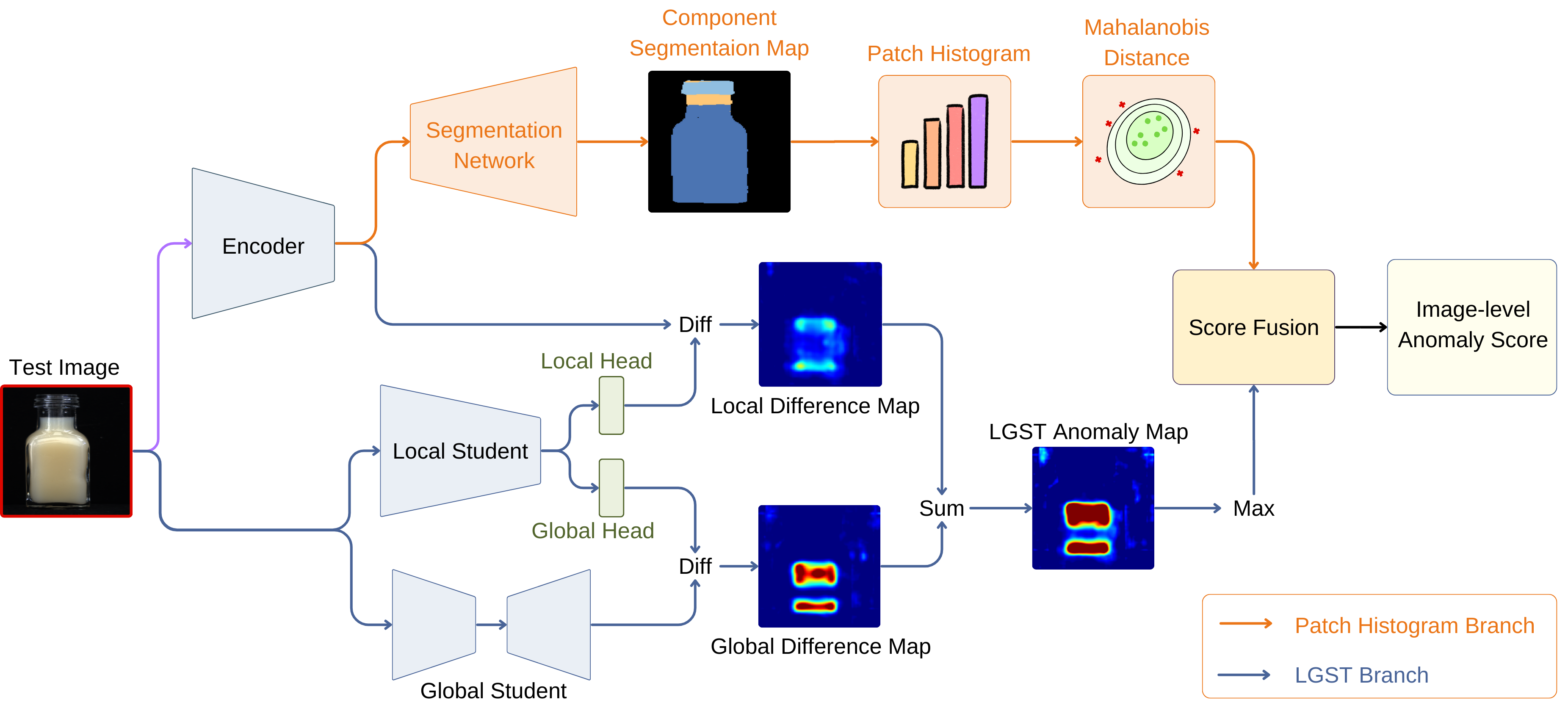}
    \caption{\textbf{Overall architecture of the proposed CSAD in the inference stage.} It consists of two branches: a Patch Histogram branch that detects anomalies using component segmentation and an LGST branch that detects both small and large-scale anomalies.}
    \label{fig:lgst}
\end{figure}

{\bf Model Architecture}\quad 
Similar to the architecture of EfficientAD~\cite{batzner2024efficientad}, our Local-Global\\ Student-Teacher(LGST) branch comprises two student models and one teacher network as shown in Figure \ref{fig:lgst}.
The student model's primary goal is to learn the normal feature distribution from the teacher. Since the students are not exposed to abnormal feature distributions, the feature output in anomalous regions will differ between the teacher and students. To detect small-scale anomalies, the local student network extracts features with a 33x33 receptive field. In contrast, the global student utilizes a bottleneck design that reduces the spatial dimension to 1x1, making the receptive field cover the entire image. This dual approach effectively captures both small-scale and large-scale anomalies.

\noindent
{\bf Difference between LGST and EfficientAD}\quad 
In the EfficientAD framework, the local student is required to output a feature map with double the channels. Half of these channels are dedicated to computing the local anomaly map, and the other half to computing the global anomaly map. In our LGST module, we avoid doubling the channels. Instead, after the local student, we employ two separate heads, each consisting of one layer of 1x1 convolution. These heads generate two feature maps, each retaining the channel size the same as the teacher's output. This modification can reduce computational cost while only slightly affecting performance.
Next, we replace the distilled PDN with a pretrained CNN, which allows us to make use of both shallow and deep features for the LGST and the segmentation network in a single forward pass, significantly reducing overall latency.

\subsection{Anomaly Score}
\label{scorefusion}
\noindent
{\bf Score Fusion}\quad
To effectively combine the image-level anomaly scores derived from both the Patch Histogram and LGST branches, a direct summation is not feasible due to differences in scale and variation. To tackle this issue, we normalize Patch Histogram anomaly scores and LGST anomaly scores separately before summation. The normalized score, denoted $ \hat{S}$, is computed using the formula $\hat{S}= (S-\mu_s)/\sigma_s $, where $S$ is the original anomaly score, $\mu_s$ and $\sigma_s$ denote the trimmed mean and the trimmed standard deviation, respectively, calculated from the anomaly scores of the validation set with the trimmed range equal to (20\%, 80\%). This range helps to filter out the extreme values. The final anomaly score is the summation of the normalized Patch Histogram anomaly score and normalized LGST anomaly score.

\begin{table*}[t]
    \begin{center}
    %% \tablesize{} %% You can specify the fontsize here, e.g., \tablesize{\footnotesize}. If commented out \small will be used.
    \resizebox{\textwidth}{!}{%
    \begin{tabular}{clccccccc}
    \multirow{2}{*}[-2pt]{\textbf{MVTec LOCO AD}}&
    \multirow{2}{*}[-2pt]{\textbf{Category}}&
    \multicolumn{4}{c}{\textbf{No Segmentation}} & \textbf{Supervised} & \multicolumn{2}{c}{\textbf{Unsupervised}} \\
    
    \cmidrule(lr){3-6}
    \cmidrule(lr){7-7}
    \cmidrule(lr){8-9}
    
    &
    &
    SimpleNet\cite{liu2023simplenet}& 
    PatchCore\cite{roth2022towards} & 
    AST\cite{rudolph2023asymmetric}    & 
    EfficientAD\cite{batzner2024efficientad}    &
    PSAD*\cite{kim2024few}& 
    ComAD\cite{LIU2023102161} & 
    \textbf{CSAD(Ours)} \\

    \hline
    \multirow{6}{*}[-2pt]{\bf{\shortstack{Logical \\Anomalies\\(LA)}}} & Breakfast Box      & 77.1  & 74.8    & 80.0    & 85.5    & \textbf{100.0}    &  91.1 & \underline{94.4}      \\
    & Juice Bottle     & 87.8  & 93.9   & 91.6    & \underline{98.4}   & \textbf{99.1}    &  95.0   & 94.9   \\
    & Pushpins        & 69.0   & 63.6   & 65.1   & 97.7    & \textbf{100.0 }  &  95.7  & \underline{99.5}    \\
    & Screw Bag        & 51.6   & 57.8    & 80.1    & 56.7    & \underline{99.3}    &  71.9  & \textbf{99.9}    \\
    
    & Splicing Connectors        & 72.0  & 79.2   & 81.8    & \textbf{95.5}   & 91.9    &  93.3  & \underline{94.8}   \\
    \cline{2-9}
    & Average(Logical)     & 71.5   & 73.9    & 79.7    & 86.8    & \textbf{98.1}    &  89.4   & \underline{96.7}    \\
    \hline
    \multirow{6}{*}[-2pt]{\bf{\shortstack{Structural \\Anomalies\\(SA)}}} & Breakfast Box      & 80.9   & 80.1    & 79.9    & \underline{88.4}    & 84.9    &  81.6  & \textbf{91.1}      \\
    & Juice Bottle     & 90.4  & \underline{98.5}   & 95.5    & \textbf{99.7}   & 98.2    &  98.2   & 95.6   \\
    & Pushpins        & 81.6   & 87.9   & 77.8   & \underline{96.1}    & 89.8   &  91.1  & \textbf{97.8}    \\
    & Screw Bag        & 83.3   & 92.0    & \textbf{95.9}    & 90.7    & \underline{95.7}    &  88.5   & 93.2    \\
    
    & Splicing Connectors        & 82.6  & 88.0   & 89.4    & \textbf{98.5}   & 89.3    &  \underline{94.9}  & 92.2   \\
    \cline{2-9}
    & Average(Structural)     & 83.7   & 89.3    & 87.7    & \textbf{94.7}    & 91.6    &  90.9   & \underline{94.0}    \\
    \hline
    \multicolumn{2}{c}{\textbf{Total Average}}       & 77.6   &  81.6   & 83.7    & 90.7    & \underline{94.8}    &  90.1   & \textbf{95.3}\\
    \hline
    \hline
    \multicolumn{2}{c}{\textbf{Throughput}}       & 114.3   &  109.9   & 28.2    & \textbf{2369.8}    & 10.2    &  69.9   & \underline{321.8}\\
    \hline
    \multicolumn{2}{c}{\textbf{Latency}}       & 19.6    &  17.3   & 58.7    & \textbf{1.9}    & 97.8    &  19.5   & \underline{8.9}\\
    \hline
    \end{tabular}}
    \end{center}
    
    \caption{Comparison of MVTec LOCO performance with state-of-the-art methods, as measured by image AUROC. The top row indicates which type of component segmentation is used. The mark * denotes that the throughput is measured with a batch size equal to 1 since PSAD does not provide a batch inference. The highest and second highest scores are highlighted in \textbf{bold} and \underline{underline}, respectively.}
    \label{tab:result1}
\end{table*}

\section{Experiments}
\subsection{Implementation Details}

\noindent
{\bf Segmentation Network}\quad
We utilize features extracted from the first and third layers of the ImageNet~\cite{deng2009imagenet} pretrained WideResNet-50~\cite{zagoruyko2016wide} as inputs for the DeepLabV3~\cite{chen2018encoder} decoder. All images are resized to 256x256.
The weights $\lambda_1, \lambda_2, \lambda_3, \lambda_4, \lambda_5$ are empirically assigned to 0.5, 10, 1, 10, and 1, respectively. Over 120 epochs, training focuses exclusively on the segmentation network decoder, with the initial 50 epochs using only labeled images.

\noindent
{\bf LGST}\quad 
Our training scheme for the LGST branch is the same as that used in EfficientAD. Images are resized to 256×256, and the feature map of one image extracted from all sub-networks in LGST has the same shape of [512,56,56].

\subsection{Experimental Setup}
We evaluate our method on the MVTec LOCO AD~\cite{bergmann2022beyond} dataset and compare its performance with SimpleNet~\cite{liu2023simplenet}, PatchCore~\cite{roth2022towards}, AST~\cite{rudolph2023asymmetric}, EfficientAD~\cite{batzner2024efficientad}, PSAD~\cite{kim2024few} and ComAD~\cite{LIU2023102161}, focusing on their detection performance, latency, and throughput. All experiments are run on a PC equipped with an RTX-3090 GPU and a Core i5-13600K CPU.

\quad Following established methods, we use AUROC as our evaluation metric.
For the speed analysis, we measure latency with a batch size of 1 and report the mean latency of 500 runs of inference. The throughput is calculated by $throughput=(batch\_size\times runs)/total\_time$ with 500 runs and a batch size of 8.

\subsection{Experimental Results}
\noindent
{\bf Performance on MVTec LOCO AD}\quad
As shown in Figure \ref{tab:result1} of the MVTec LOCO benchmark, our method achieves an AUROC of 94.0\% for structural anomalies and 96.7\% for logical anomalies, showcasing superior detection capabilities. With a remarkable total average AUROC of 95.3\%, our approach excels in achieving high accuracy on both logical and structural anomalies simultaneously.

\noindent
{\bf Latency and Throughput}\quad Table \ref{tab:result1} presents a speed comparison between our method and other approaches. Although our method does not surpass the impressive latency and throughput of EfficientAD, the latency of our method is just 8.9 milliseconds, demonstrating a considerable improvement over other segmentation-based models like PSAD and ComAD. The throughput of our method is also noteworthy at 321.8 images per second, indicating a competitive capability for high-speed anomaly detection. Figure \ref{fig:speed} is the speed-performance plot of our and other SOTA methods. It shows that our method reaches a new SOTA performance with the highest anomaly detection accuracy and lower latency than most existing approaches.

\begin{table}[h]
\vspace{5pt}
\captionsetup{width=\linewidth}
\centering
\resizebox{0.5\textwidth}{!}{
    \begin{tabular}{p{3cm}ccc}
    \toprule
    \textbf{Model} & \textbf{LA} & \textbf{SA} & \textbf{Mean} \\
    \midrule
    PSAD & \textbf{94.2} & 71.1 & \underline{82.7}\\
    ComAD & 87.7 & \underline{74.6} & 81.2\\
    PatchHist& \underline{91.4} & \textbf{75.4} & \textbf{83.4}\\
    \bottomrule
    \end{tabular}
    }
    \caption{MVTec LOCO AD performance comparison of segmentation modules of different methods and different settings measured in AUROC. LA and SA denote logical anomalies and structural anomalies, respectively.}
    \label{tab:seg}
    \vspace{-12pt}
\end{table}

\begin{table}[h]
\vspace{10pt}
\captionsetup{width=\linewidth}
\centering
\resizebox{0.5\textwidth}{!}{
    \begin{tabular}{p{3cm}ccc}
    \toprule
    \textbf{Augmentation} & \textbf{LA} & \textbf{SA} & \textbf{Mean} \\
    \midrule
    None & \underline{89.2} & \underline{71.0} & \underline{80.1}\\
    CutPaste & 89.0 & 70.1 & 79.6\\
   
    LSA & \textbf{89.9} & \textbf{73.7} & \textbf{81.8}\\
    
    \bottomrule
    \end{tabular}
    }
    \caption{MVTec LOCO AD performance comparison of different augmentation methods used in training the segmentation network. The detection performance is measured in AUROC.}
    \label{tab:aug}
\end{table}

\noindent
{\bf Segmentation Branch}\quad In Table \ref{tab:seg}, we compare our Patch Histogram branch with other segmentation-based methods. We only compare the segmentation-related modules. In detail, we report the score of ComAD excludes the PatchCore branch and the score of PSAD with only the histogram memory bank; the scores are retrieved from their original papers. As for our method, we report the score using only the Patch Histogram branch.
PSAD attains the highest scores in detecting logical anomalies through precise human annotations. Meanwhile, our Patch Histogram has achieved an AUROC of 75.4\% in structural anomalies, outperforming other techniques and securing the highest average score of 83.4\% AUROC. These results prove the effectiveness of our Patch Histogram design in detecting both types of anomalies.

{\bf Augmentation of Semantic Pseudo-label Map}
\quad Table \ref{tab:aug} shows the comparison of using LSA and CutPaste~\cite{li2021cutpaste} during training the segmentation network. The result demonstrates that our LSA augmentation improves the segmentation network's robustness. Unlike CutPaste, which randomly cuts and pastes image patches, LSA augments images and labels while preserving the semantic integrity of the components. For instance, in the "screw bag" category, the model can not distinguish between long screws and short screws with just partial screws in CutPaste-augmented images, thereby impairing anomaly detection performance.

\subsection{Ablation Study}
\noindent
{\bf Patch Size of Patch Histogram}\quad
Table \ref {tab:patchsize} illustrates that each patch size combination in the histogram delivers consistently similar performance across various settings, significantly improving the detection of both logical and structural anomalies compared to that without using patch histogram (patch size=256). However, as the patch size decreases,
there is a notable increase in latency. Considering latency and performance, we select 256+128 as our setting. The score fusion for different patch sizes is the same as that described in Sec \ref{scorefusion}.

\noindent
{\bf Impact of Different Components and Settings of CSAD}\quad Table \ref{tab:modules} shows the performance and speed of different settings of CSAD. From the first three rows, we compared our local student design with that of EfficientAD, the result shows that our design can reduce latency with a small performance drop. From the fourth and fifth rows, LSA improves both logical and structural detection performance. The last two rows indicate that using either the class histogram or the patch histogram results in a comparable logical anomaly detection in CSAD, we attribute this to the limitation as noted in the supplementary material. In contrast, using patch histograms can improve the detection performance of structural anomalies due to the component area variation in some structural anomalies.

\begin{table}[h]
\centering
\captionsetup{width=\linewidth}
\resizebox{0.6\textwidth}{!}{
    \begin{tabular}{cccccccc}
    \toprule
    \multicolumn{4}{c}{\textbf{Patch Size}}&
    \multirow{2}{*}[-2pt]{\textbf{LA}}&
    \multirow{2}{*}[-2pt]{\textbf{SA}} & \multirow{2}{*}[-2pt]{\textbf{Mean}} & \multirow{2}{*}[-2pt]{\textbf{Latency(ms)}}\\
    256&128&85&64&&&&\\
    \midrule
    \checkmark &&& & 89.9 & 73.7 & 81.8 & \textbf{4.2}\\
    \checkmark &\checkmark && & \underline{91.4} & 75.4 & \underline{83.4} & \underline{5.6}\\
    \checkmark &\checkmark &\checkmark & & \textbf{91.6} & \textbf{76.0} & \textbf{83.8} & 10.1\\
    \checkmark &\checkmark &&\checkmark & 90.8 & 75.2 & 83.0 & 9.8\\
    \checkmark &\checkmark &\checkmark &\checkmark & 91.1 & \underline{75.7} & \underline{83.4} & 14.1\\
    
    \bottomrule
    \end{tabular}
    }
    \caption{Performance and speed of different patch size combinations in the Patch Histogram module. The detection performance is measured in AUROC.}
    \label{tab:patchsize}
\end{table}

\begin{table}[h]
\centering
\vspace{-5pt}
\captionsetup{width=\linewidth}
    \resizebox{0.8\textwidth}{!}{
    \begin{tabular}{ccccccc}
    \toprule
    \multicolumn{3}{c}{\textbf{Setting}}&
    \multirow{2}{*}[-2pt]{\textbf{LA}}&
    \multirow{2}{*}[-2pt]{\textbf{SA}} & 
    \multirow{2}{*}[-2pt]{\textbf{Mean}} & 
    \multirow{2}{*}[-2pt]{\textbf{Latency(ms)} }\\
    
    Local Student&PatchHist&LSA&&\\
    
    \midrule
     Same&\ding{56}&\ding{56}&72.5&63.5&68.06&\textbf{6.51}\\
    
    DoubleChannel&\ding{56}&\ding{56}&88.0&92.8&90.40&\underline{6.89}\\
    
    Heads&\ding{56}&\ding{56}&87.7&93.1&90.38&6.73\\
    
    Heads&256&\ding{56}&\underline{96.5}&92.4&94.45&7.84\\
    Heads&256&\checkmark&\textbf{96.7}&\underline{93.8}&\underline{95.25}&7.84\\
    Heads&256+128&\checkmark&\textbf{96.7}&\textbf{94.0}&\textbf{95.34}&8.96\\
    
    \bottomrule
    \end{tabular}
    }
    \caption{Impact of different components and settings of CSAD. In the design of the local student, "Same" means that the output feature maps of the local student used to calculate local and global difference maps are the same, "DoubleChannel" is the design of EfficientAD, and "Heads" is the design of CSAD. The detection performance is measured in AUROC.}
    \label{tab:modules}
\end{table}

\section{Conclusion}

In this paper, we presented a segmentation-based approach that significantly improves industrial anomaly detection. We achieved accurate component segmentation without human annotations by exploiting multiple foundation models. Integrating the component segmentation network with the Patch Histogram and LGST modules, our method outperforms the current state-of-the-art methods with higher accuracy and lower latency, as demonstrated in our experiments on the MVTec LOCO AD dataset.\\

\vspace{10pt}
{\Large \color{bmv@sectioncolor} \bf Acknowledgments}

This work was supported in part by the National Science and Technology Council, Taiwan under grants NSTC 111-2221-E-007-106-MY3 and NSTC 112-2634-F-007 002.

\bibliography{egbib}

\newpage

\setcounter{section}{0}
\renewcommand{\thesection}{\Alph{section}}

\section{Semantic Pseudo-label Generation}
\subsection{Image Tag Generation}

Table \ref{tab:tags} shows the image tags of all categories generated by RAM++~\cite{huang2023open} and the tags after manual filtering. We use the first image in the training set to generate the tags; for categories containing multiple types of products, we generate tags for each type and merge them together.

\subsection{Component Mask Generation}

In the component mask generation, we use the pretrained weight \textit{sam\_hq\_vit\_h.pth} from SAM-HQ~\cite{sam_hq}, a variation of SAM trained on high-quality datasets, as our weight in SAM.
We use the pretrained weight \textit{grounding-dino-swinT-OGC.pth} and the same \textit{sam\_hq\_vit\_h.pth} as our weights in Grounded-SAM.

\subsection{Mask Refinement}
In the mask refinement process, we use two algorithms to obtain the refined mask from $X^{sem}$ and $X^{seg}$, the filter-by-grounding algorithm and the filter-by-combine algorithm. The filter-by-combine algorithm is used to filter out overlapped masks, while filter-by-grounding algorithm is used to filter out noise masks. For the "screw bag" category, we apply both the filter-by-grounding algorithm and the filter-by-combine algorithm. As for the "juice bottle," we only apply the filter-by-combine algorithm. Detailed description of these two algorithms is given below.

\noindent
{\bf Filter-by-grounding}\quad
The filter-by-grounding algorithm filters out noise masks by removing masks in $X^{seg}$ that are not highly overlapped with $X^{sem}$. The Python code using Numpy is shown in Figure \ref{alg:grounding}

\noindent
{\bf Filter-by-combine}\quad
In the masks SAM generates, some overlapped masks need to be processed. For example, in the "screw bag" category, the SAM generates not only the masks of the head and body of the screw but also the mask of the whole screw. In this case, we aim to filter out the mask of the whole screw since we want precise component segmentation at this stage. The filter-by-combine algorithm checks if one mask can be the combination of some small masks. If so, this mask is dropped. After the process, we can remove the masks that are highly overlapped with other masks and maintain a fine-grained segmentation map. The Python code using Numpy is shown in Figure \ref{alg:combine}.

\subsection{Component Feature Extraction}
We use ImageNet pretrained WideResNet-50 as our feature extractor. The feature maps are obtained from the fourth layer of the feature extractor, and the rotation augmentation of each component is set to 60.

\subsection{Resulting Semantic Pseudo-label Maps}
Figure \ref{fig:pseudo-label} shows the example images of $X^{seg}$, $X^{sem}$ and the pseudo-label maps, as described in Sec.\ref{limitation}. There are components missing in the pseudo-label maps in the category "breakfast box" and "juice bottle", and we fill up the holes for each component to mitigate this issue.

\section{Component Segmentation}
\subsection{Training Process of Segmentation Network}
\begin{figure}[!ht]
    \centering
    \includegraphics[width=0.8\textwidth]{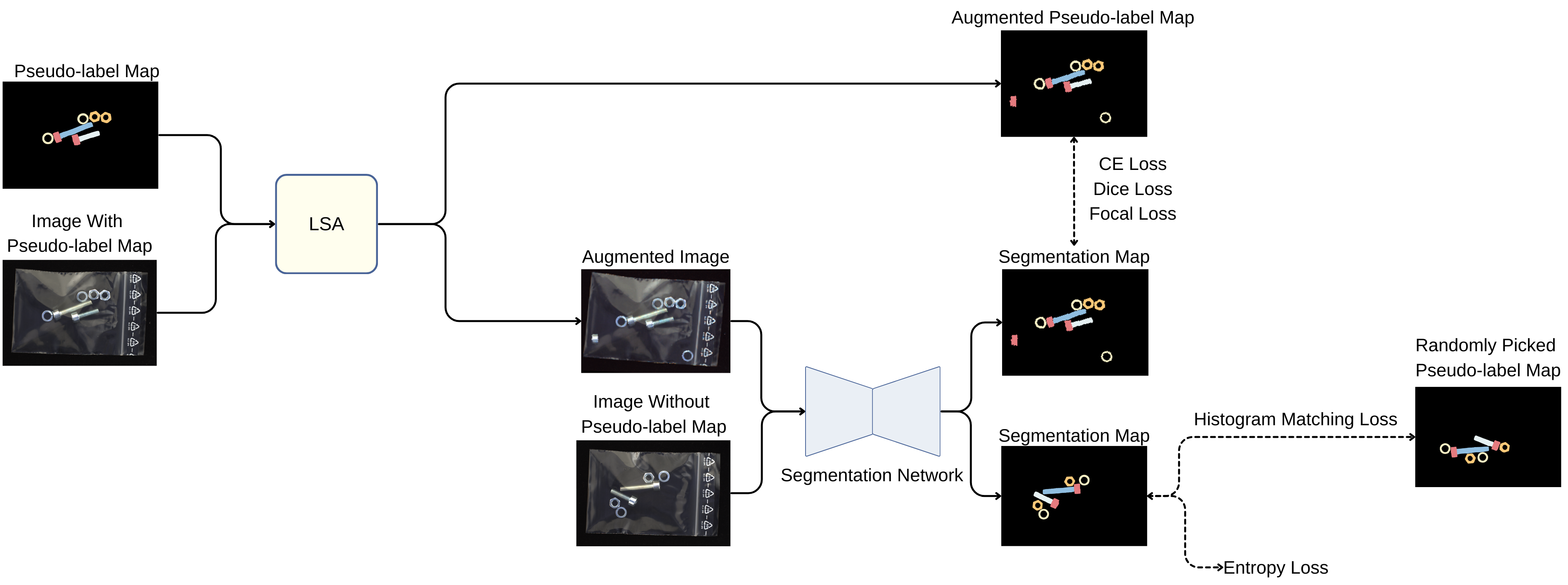}
    \caption{Diagram of the training process of the segmentation network.}
    \label{fig:seg}
\end{figure}

Figure \ref{fig:seg} shows the training process of the segmentation network. All images with semantic pseudo-label maps are augmented through LSA and the predictions of augmented images and augmented pseudo-label maps are used to calculate Cross Entropy loss, Dice loss, and Focal loss. Predictions of unlabeled images and randomly picked pseudo-label maps are used to calculate Histogram Matching loss, another Entropy loss that calculates the average entropy of each pixel's prediction is applied to reduce the uncertainty of the prediction.

\subsection{LSA Augmentation}

\begin{figure}[!ht]
    \centering
    \includegraphics[width=0.8\textwidth]{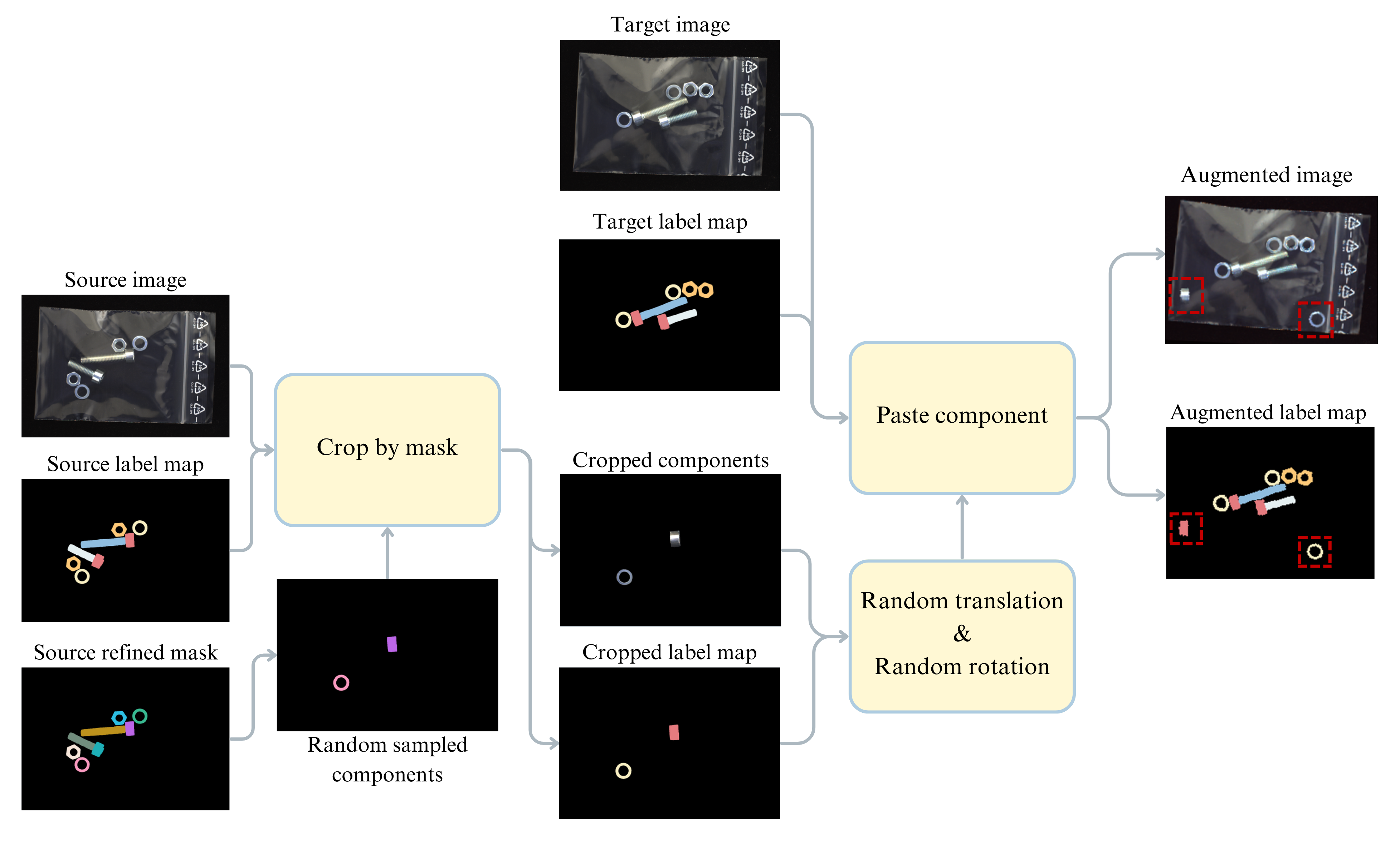}
    \caption{Diagram of the LSA augmentation process. The source image is randomly sampled from the images used in supervised training.}
    \label{fig:lsa}
\end{figure}

\begin{figure}[!ht]
    \centering
    \includegraphics[width=0.6\textwidth]{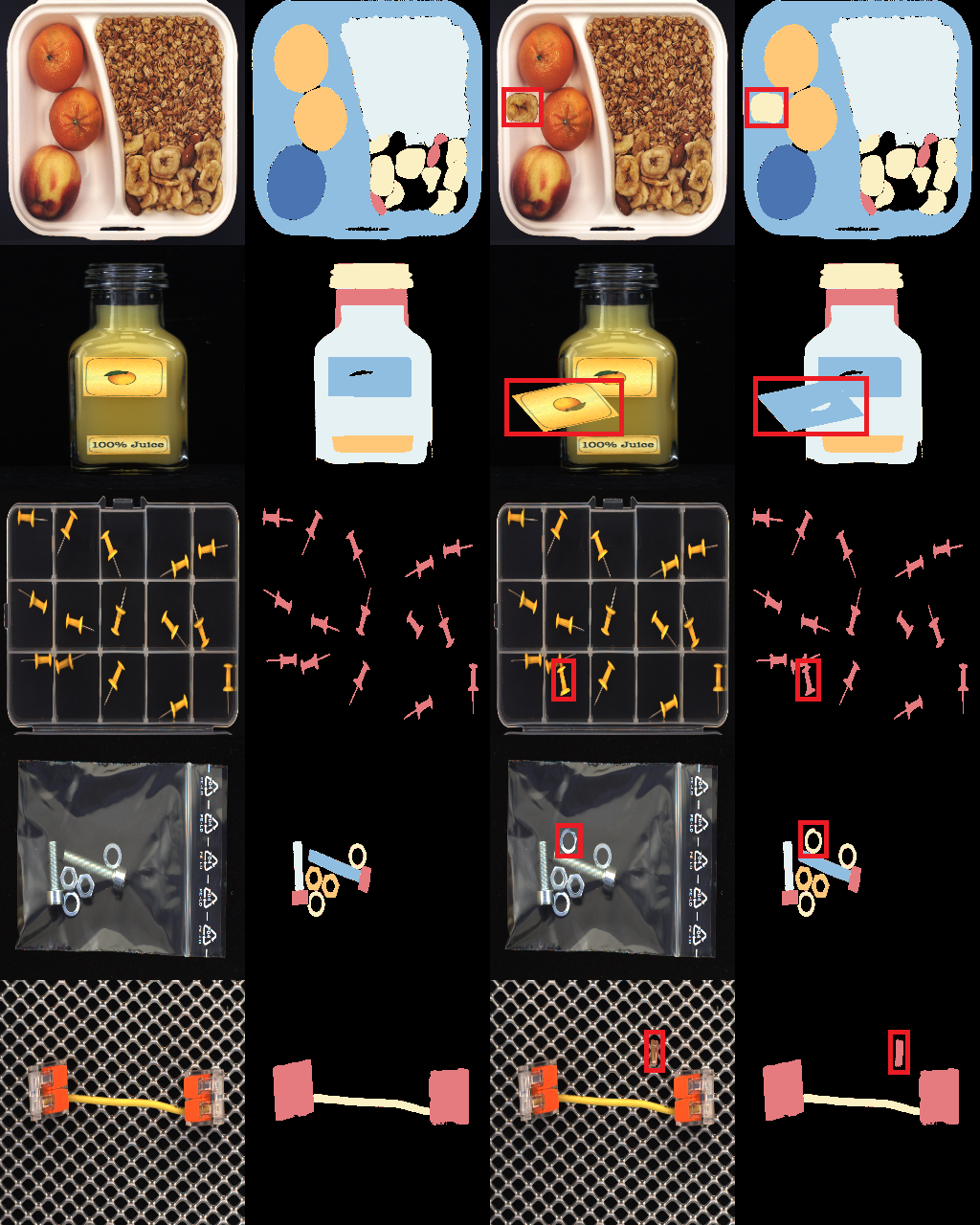}
    \caption{Example images of the LSA. From left to right presents the original image, pseudo-label map, LSA augmented image, and LSA augmented pseudo-label map. The red bounding box indicates the additional component added to the image.}
    \label{fig:vislsa}
\end{figure}

%\subsection{LSA}
Figure \ref{fig:lsa} shows the proposed LSA augmentation procedure. To simulate logical anomalies, the position of the additional component is set at least $0.1\times image\_size$ far from the original position. Figure \ref{fig:vislsa} shows the original images, the pseudo-label maps, the images augmented by LSA, and the augmented pseudo-label maps. 

\section{Implementation Detail}
%\subsection{Implementation Detail}
The performance scores of the compared methods in all our experiments are retrieved from the original papers, except that the score of PatchCore~\cite{roth2022towards} is from the paper of PSAD~\cite{kim2024few} and the score of SimpleNet~\cite{liu2023simplenet} is from the paper of EfficientAD~\cite{batzner2024efficientad}.
In the speed comparison section, we use the official implementation except for the EfficientAD, which does not have an official implementation. Instead, we use the implementation of Anomalib\cite{akcay2022anomalib}.

\section{Anomaly Localization}
\begin{figure}[t]
    \centering
    \includegraphics[width=0.95\textwidth]{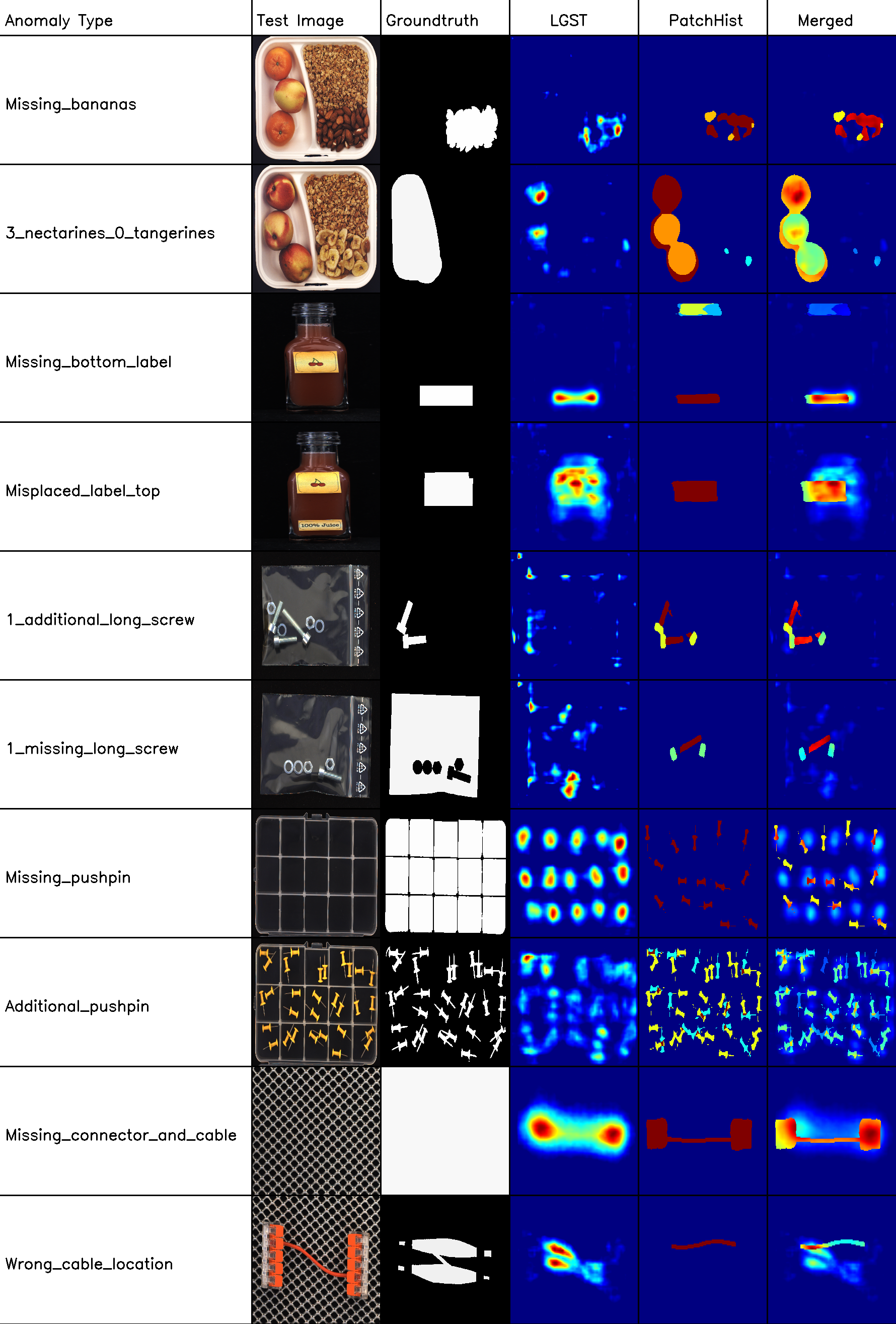}
    \caption{Example images and anomaly maps of anomaly localization result. For each category, we show two types of different logical anomalies.}
    \label{fig:anomalymap}
\end{figure}

To obtain anomaly localization results, we merge the anomaly maps from the LGST and Patch Histogram branches. The LGST anomaly maps are generated by averaging the difference maps of local and global student networks. For the Patch Histogram branch, we analyze the difference between the current histogram and the mean histogram of the training set. This analysis identifies missing or additional component classes. For additional components, we assign the class region the anomaly score based on the difference of the histogram. For missing components, we search for a normal image in the training set with the closest class histogram and assign the anomaly score based on the difference to that class region of the normal image. The final patch histogram anomaly map is the sum of all anomaly maps of different patch sizes. The example images and the corresponding anomaly maps are shown in Figure \ref{fig:anomalymap}.

\quad Note that the latency of anomaly map generation is not included in the experiments since it does not affect the image level detection result.

\subsection{Anomaly Localization Performance}
The anomaly localization performance is shown in Table \ref{tab:spro}. While our method does not outperform EfficientAD due to the limitations outlined in Sec. \ref{limitation} , our Patch Histogram branch improves anomaly localization performance in both logical and structural anomalies, with an average sPRO improvement of 1.0\%.

\begin{table}[!h]
\captionsetup{width=\linewidth}
\centering
\resizebox{\textwidth}{!}{
    \begin{tabular}{lcccccc}
    \toprule
    \textbf{Model} & 
    Breakfast Box &
    Juice Bottle &
    Pushpins &
    Screw Bag &
    Splicing Connectors & 
    Average\\
    \midrule
    GCAD & - & - & - & - & - & 89.1 \\
    EfficientAD & - & - & - & - & - & 92.5 \\
    LGST & 83.2 & 95.2 & 87.8 & 81.1 & 94.4 & 88.3 \\
    CSAD & 84.0 & 95.3 & 88.1 & 84.6 & 94.5 & 89.3 \\
    \midrule
    \midrule
    LGST & 76.7 / 89.6 & 95.2 / 95.3 & 98.5 / 77.1 & 73.9 / 88.3 & 94.9 / 93.9 & 87.8 / 88.8 \\
    CSAD & 78.1 / 89.9 & 91.7 / 98.8 & 99.0 / 77.2 & 82.1 / 87.1 & 94.8 / 94.1 & 89.1 / 89.4 \\
    \bottomrule
    \end{tabular}
    }

    \caption{Anomaly localization performance of the MVTec LOCO AD dataset. The result is reported in sPRO. For the last two rows, the score is given by (logical sPRO / structural sPRO).}
    \label{tab:spro}
\end{table}

\section{Supplementary Experiments}
\subsection{Performance and Speed of Different Branches in
CSAD}
Table \ref{tab:LGST} shows the anomaly detection performance and the speed of different branches in our method. The results indicate that while the LGST and Patch Histogram branches individually do not achieve optimal performance, their combination yields the highest scores for both logical and structural anomalies. Remarkably, the Patch Histogram branch enhances structural anomaly detection despite not being specifically designed for it. Furthermore, LGST and Patch Histogram exhibit latencies of 5.7 milliseconds and 5.6 milliseconds, respectively. The overall latency of CSAD is reduced to 8.9 milliseconds, with a 2.4-millisecond reduction attributed to the reuse of features extracted by the teacher.

\begin{table}[!h]
\centering
\vspace{20pt}
\captionsetup{width=.8\linewidth}
\resizebox{0.8\textwidth}{!}{
    \begin{tabular}{ccccccc}
    \toprule
    \multicolumn{2}{c}{\textbf{Branch}}&
    \multirow{2}{*}[-2pt]{\textbf{LA}}&
    \multirow{2}{*}[-2pt]{\textbf{SA}} & 
    \multirow{2}{*}[-2pt]{\textbf{Mean}} & 
    \multirow{2}{*}[-2pt]{\textbf{Latency(ms)} }&
    \multirow{2}{*}[-2pt]{\textbf{Throughput(fps)}}\\
    PatchHist&LGST&&&&&\\
    \midrule
    
    \checkmark& & \underline{91.4} & 75.4 & 83.4 & \textbf{5.6}  & \underline{603.5}\\
    &\checkmark & 87.7 & \underline{93.1} & \underline{90.4} & \underline{5.7}  & \textbf{1402.9}\\
    \checkmark&\checkmark & \textbf{96.7} & \textbf{94.0} & \textbf{95.3} & 8.9 & 321.8\\
    
    \bottomrule
    \end{tabular}
    }
    \caption{Anomaly detection performance and speed comparison of different branches in our method. LA and SA denote logical anomalies and structural anomalies, respectively. Anomaly detection performance is measured in AUROC.}
    \label{tab:LGST}
\end{table}

\subsection{Hyperparameter of Component Clustering}
\begin{figure}[!ht]
    \centering
    \includegraphics[width=0.95\textwidth]{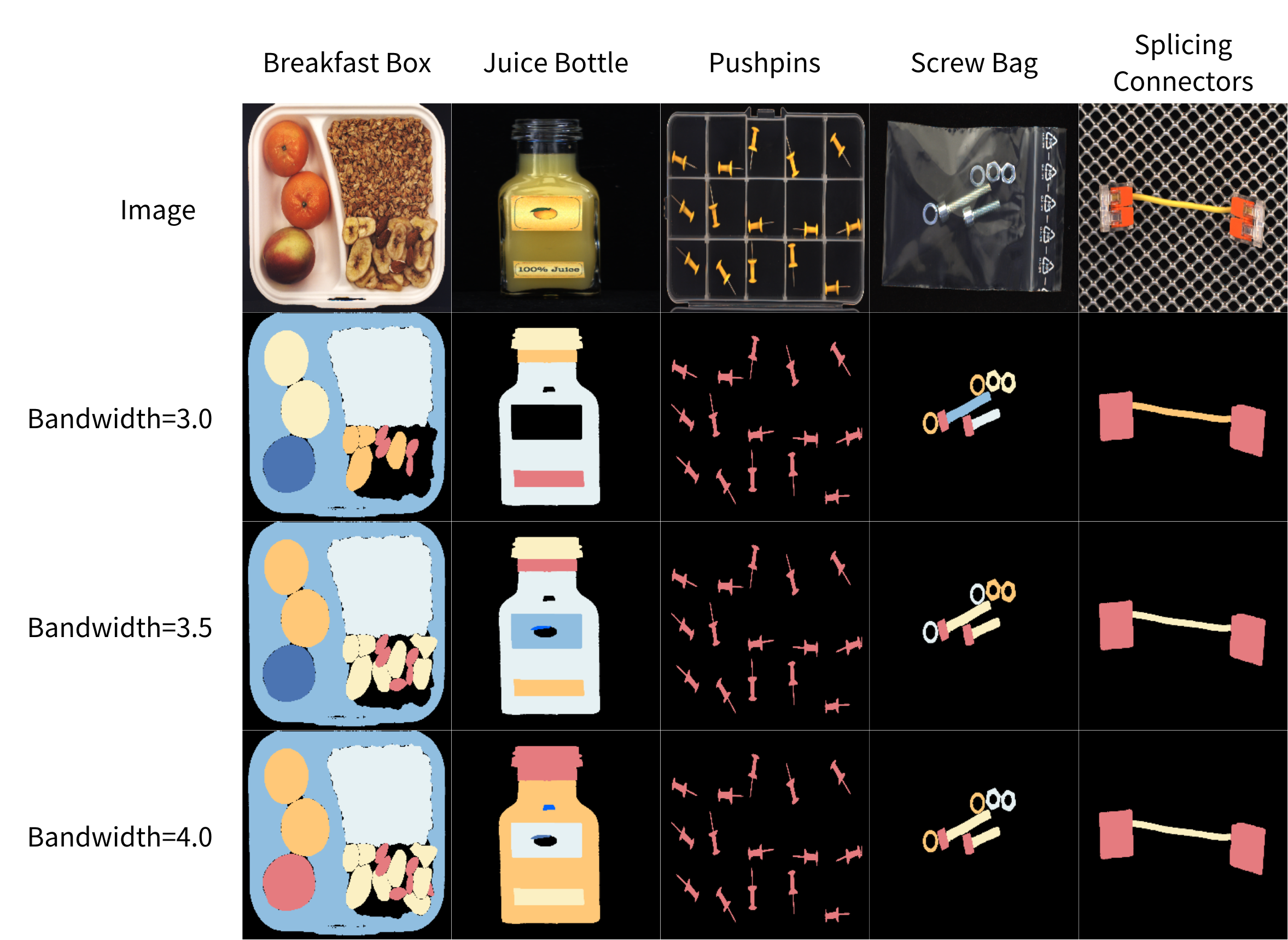}
    \caption{Result of component clustering under different hyperparameters. For each row, the same color indicates the same cluster of components.}
    \label{fig:cluster}
\end{figure}

Figure \ref{fig:cluster} illustrates the result of component clustering under different hyperparameters. Three bandwidths were tested in MeanShift clustering. A larger bandwidth results in less similar components being clustered together, potentially causing coarse semantic segmentation. For instance, a bandwidth of 4.0 clusters both long and short screw bodies together, which should be separated. In our experiments, We select bandwidth=3 for "screw bag" and 3.5 for all other categories.

\section{Limitation}

\subsection{Limitation of Patch Histogram}
The limitation of our patch histogram lies in the selection of patch sizes. Categories with strict component position constraints, such as "splicing connectors," benefit from smaller patch sizes, while those without such constraints may be adversely affected like "screw bag." Therefore, selecting the appropriate patch size requires prior knowledge of the product and its specific application scenarios.

\subsection{Limitation of Semantic Pseudo-label Generation}
\label{limitation}
\begin{figure}[h]
\centering
  \includegraphics[width=0.55\textwidth]{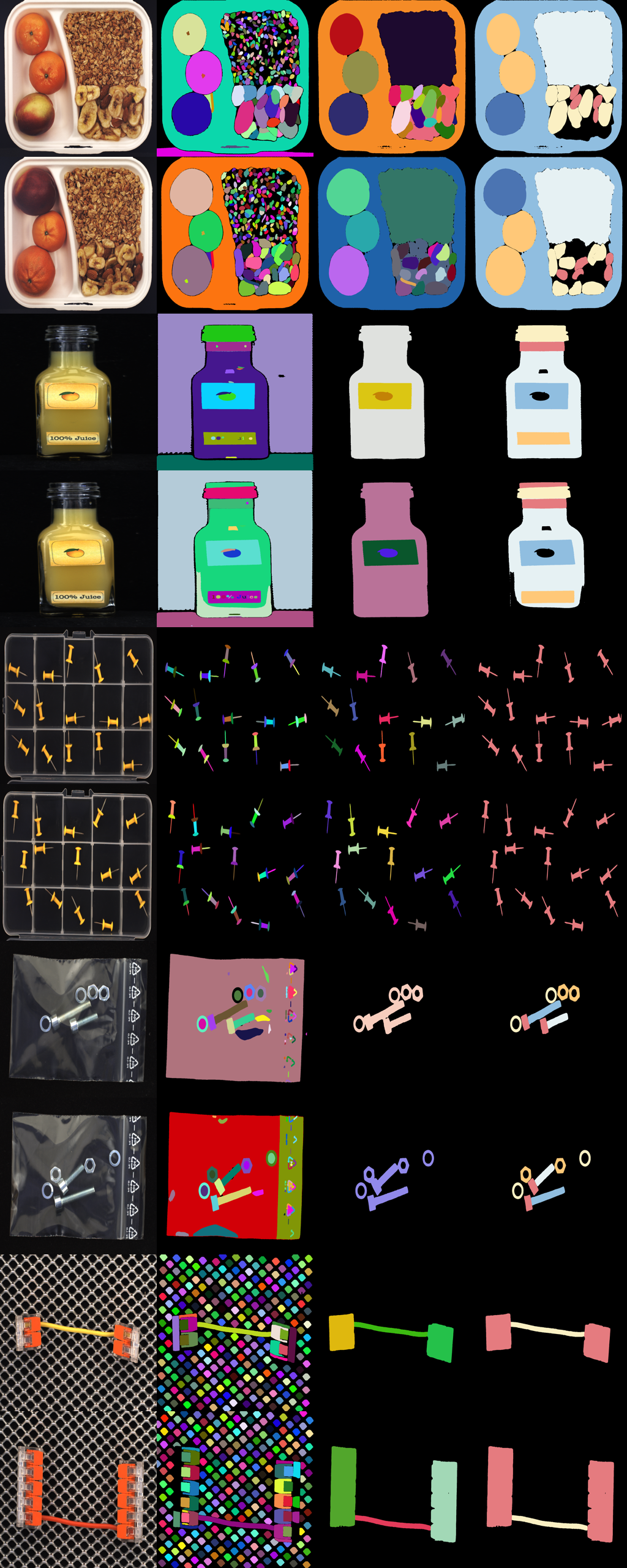}
  \caption{
  Example images of pseudo label generation of the five categories of the MVTecc LOCO, from left to right, represent normal image, $X^{seg}$, $X^{sem}$, and pseudo label image. In each category, random colors are assigned to different masks in $X^{seg}$ and $X^{sem}$ while the same color represents the same class in the pseudo-label image.
    }
  \label{fig:pseudo-label}
\end{figure}

Our pseudo-label generation process generally produces precise label maps. However, it is limited by the performance of the foundational model used in our implementation. This limitation was evident when the model failed to accurately segment obscured almonds in the "breakfast box," causing these regions to be incorrectly classified as background. Another example is the fruit icon missing in the "juice bottle", which is filtered out in the component clustering stage since the number of each unique fruit icon did not exceed half of the training samples, as described in Section 3.1 of the paper. Example images are shown in Figure \ref{fig:pseudo-label}. Despite this limitation, our Patch Histogram module is still capable of identifying logical anomalies that arise from such segmentation errors. Future efforts could focus on improving segmentation in such challenging scenarios.

\begin{table}[!ht]
    \begin{tabular}{|c|c|c|}
    \hline
    Category & RAM++ Output Tags & Filtered Tags\\
    \hline
    Breakfast Box & \shortstack{ almond, apple, banana,\\ cereal, container, fill,\\ food, fruit, grain,\\ granola, mixture, nut,\\ oatmeal, oats, orange,\\ plastic, raisin, seed,\\ topping, tray} & \shortstack{almond, apple, container,\\ oatmeal, orange, banana} \\
    \hline
    Juice Bottle & \shortstack{alcohol, apple juice,\\ beverage, bottle, liquor,\\ glass bottle, juice,\\ lemonade, liquid, olive oil,\\ orange juice, yellow, alcohol, \\banana, bottle, glass bottle,\\ glass jar, jug, juice,\\ liquid, milk, alcohol, beverage,\\ bottle, cherry, condiment, \\liquor, glass bottle, honey, juice, \\liquid, maple syrup, sauce, syrup,\\ tomato sauce } & \shortstack{alcohol, apple juice, beverage,\\ bottle, liquor, glass bottle,\\ juice, lemonade, liquid,\\ olive oil, orange juice,\\ yellow banana, bottle,\\ glass bottle, glass jar,\\ jug, juice, liquid,\\ milk beverage, bottle, cherry, \\condiment, liquor, glass bottle,\\ honey, juice,\\ liquid, maple syrup, sauce,\\ syrup, tomato sauce.} \\
    \hline
    Pushpins & \shortstack{box, case, container,\\ fill, needle, pin,\\ plastic, screw, tool,\\ tray, yellow} & \shortstack{pin, pushpin, drawing pin} \\
    \hline
    Screw Bag & \shortstack{bag, bolt, container,\\ nut, package, plastic,\\ screw, tool, metal bolts,\\ metal hex nuts, metal washers,\\ metal screws, zip-lock bag, screw,\\ ring} & \shortstack{metal bolts, metal hex nuts,\\ metal washers, metal screws,\\ zip-lock bag, screw,\\ ring, bag, bolt,\\ container, nut, tool} \\
    
    \hline
    \shortstack{Splicing\\ Connectors} & \shortstack{attach, cable, connect,\\ connector, hook, electric outlet,\\ plug, pole, socket,\\ wire} & \shortstack{cable, connector, hook,\\ electric outlet, plug,\\ pole, socket, wire} \\
    \hline
    \end{tabular}
    \caption{Image tags of all categories.}
    \label{tab:tags}
\end{table}

\begin{figure}[h]
\begin{tcolorbox}[title=Python code of filter-by-grounding algorithm.,colback=white,colframe=black,colbacktitle=white,coltitle=black]
\begin{lstlisting}[language=Python]
import numpy as np
def filter_masks_by_grounding(grounding_mask,masks):
    """
        grounding_mask: binary mask
        masks: list of N binary masks
    """
    new_mask = list()
    for mask in masks:
        if np.sum(np.logical_and(grounding_mask,mask))/np.sum(mask!=0) > 0.9:
            new_mask.append(mask)
    return new_mask

\end{lstlisting}

\end{tcolorbox}
\caption{Python code of the algorithm filter-by-grounding.}
\label{alg:grounding}
\end{figure}

\begin{figure}[h]

\begin{tcolorbox}[title=Python code of  filter-by-combine algorithm.,colback=white,colframe=black,colbacktitle=white,coltitle=black]
\begin{lstlisting}[language=Python]
import numpy as np
def intersect_ratio(mask1,mask2):
    intersection = np.logical_and(mask1,mask2)
    if intersection.sum() == 0:
        return 0
    ratio = np.sum(intersection)/min([np.sum(mask1!=0),np.sum(mask2!=0)])
    ratio = 0 if np.isnan(ratio) else ratio
    return ratio

def filter_by_combine(masks):
    """
        masks: list of N binary masks
    """

    masks = sorted(masks,key=lambda x:np.sum(x)) # small to large
    combine_masks = np.zeros_like(masks[0])
    result_masks = list()
    wait_masks = list()
    for i,mask in enumerate(masks):
        if intersect_ratio(combine_masks,mask) < 0.9 or i == 0:
            combine_masks = np.logical_or(combine_masks,mask)
            result_masks.append(mask)
        else:
            wait_masks.append(mask)
            
    # second chance
    if len(wait_masks) != 0:
        for mask in wait_masks:
            ratio = np.sum(np.logical_and(combine_masks,mask))/np.sum(mask!=0)
            if ratio < 0.9:
                combine_masks = np.logical_or(combine_masks,mask)
                result_masks.append(mask)
    return result_masks
\end{lstlisting}

\end{tcolorbox}
\caption{Python code of the algorithm filter-by-combine.}
\label{alg:combine}
\end{figure}

\end{document}